\documentclass{article}
\usepackage{spconf,amsmath,graphicx}
\usepackage[utf8]{inputenc} 
\usepackage[T1]{fontenc}    
\usepackage{hyperref}       
\usepackage{url}            
\usepackage{booktabs}       
\usepackage{amsfonts}       
\usepackage{nicefrac}       
\usepackage{microtype}      
\usepackage{graphicx}
\usepackage{threeparttable}
\usepackage[vlined,boxed,commentsnumbered,ruled,linesnumbered]{algorithm2e}
\usepackage{amsmath,amssymb}
\usepackage{multirow}
\usepackage{subfigure}
\usepackage{comment}
\usepackage{tablefootnote}
\usepackage{stfloats}
\usepackage{color}

\usepackage[linesnumbered, ruled]{algorithm2e}
\usepackage{bm}

\usepackage{colortbl}

\DeclareUrlCommand\url{\color{blue}}

\title{Ternary Weight Networks}
\name{Fengfu Li$^{1\dag}$, Bin Liu$^{2\dag}$,  Xiaoxing Wang$^{2}$, Bo Zhang$^{1*}$, Junchi Yan$^{2*}$\thanks{$\dag$: Equal contribution. $^{*}$ Correspondence authors.}}
\address{$^{1}$Institute of Applied Math., AMSS, CAS, Beijing, China\\
lifengfu12@mails.ucas.ac.cn, b.zhang@amt.ac.cn\\ 
$^{2}$MOE Key Lab of Artificial Intelligence, Shanghai Jiao Tong University, Shanghai, China\\	\{binliu\_sjtu, figure1\_wxx, yanjunchi\}@sjtu.edu.cn}

\begin{document}
%
\maketitle

\begin{abstract}
We present a memory and computation efficient ternary weight networks (TWNs) - with weights constrained to +1, 0 and -1. The Euclidian distance between full (float or double) precision weights and the ternary weights along with a scaling factor is minimized in training stage. Besides, a threshold-based ternary function is optimized to get an approximated solution which can be fast and easily computed. TWNs have shown better expressive abilities than binary precision counterparts. Meanwhile, TWNs achieve up to 16$\times$ model compression rate and need fewer multiplications compared with the float32 precision counterparts. Extensive experiments on MNIST, CIFAR-10, and ImageNet datasets show that the TWNs achieve much better  result than the Binary-Weight-Networks (BWNs) and the classification performance on MNIST and CIFAR-10 is very close to the full precision networks. We also verify our method on object detection task and show that TWNs significantly outperforms BWN by more than 10\% mAP on PASCAL VOC dataset. The pytorch version of source code is available at: https://github.com/Thinklab-SJTU/twns.

\end{abstract}
%
%

\section{Introduction and Related Work}
\label{sec:intro}
\vspace{-5pt}
Deep neural networks (DNN) have made significant improvements in lots of computer vision tasks such as object recognition~\cite{He2015,Krizhevsky2012,Simonyan2014,Szegedy2015} and object detection~\cite{Liuw2015,Ren2015}. This motivates interests to deploy the state-of-the-art DNN models to real world applications like smart phones, wearable embedded devices or other edge computing devices. However, these models often need considerable storage and computational power~\cite{Rastegari2016}, and can easily overburden the limited storage, battery power, and computer capabilities of the smart wearable embedded devices. As a result, it remains a challenge for the deployment.

To mitigate the storage and computational problem~\cite{Esser_2016,song_2015}, methods that seek to binarize weights or activations in DNN models have been proposed. BinaryConnect~\cite{courbariaux2015binaryconnect} uses a single sign function to binarize the weights. Binary Weight Networks~\cite{Rastegari2016} adopts the same binarization function but adds an extra scaling factor. The extensions of the previous methods are BinaryNet~\cite{Hubara2016} and XNOR-Net~\cite{Rastegari2016} where both weights and activations are binary-valued. These models eliminate most of the multiplications in the forward and backward propagations, and thus own the potential of gaining significant benefits with specialized deep learning (DL) hardware by replacing many multiply-accumulate operations by simple accumulation~\cite{Lin2015}. Besides, binary weight networks achieve up to 32$\times$ model compression rate.
Despite the binary techniques, some other compression methods focus on identifying models with few parameters while preserving accuracy by compressing existing state-of-the-art DNN models in a lossy way. SqueezeNet~\cite{Iandola2016} is such a model that has 50$\times$ fewer parameters than AlexNet~\cite{Krizhevsky2012} but maintains AlexNet-level accuracy on ImageNet.
MobileNet~\cite{mobilenet} and ShuffleNet~\cite{shufflenet} propose lightweight architectures to reduce the parameters and computation cost. Other methods propose to search efficient architectures and achieves great performance on both classification~\cite{darts,mergenas} and object detection~\cite{eautodet}. 
Deep Compression~\cite{song_2015} is another most recently proposed method that uses pruning, trained quantization and huffman coding for compressing neural networks. It reduced the storage requirement of AlexNet and VGG-16~\cite{Simonyan2014} by 35$\times$ and 49$\times$, respectively, without loss of accuracy. \textbf{This paper has the following contributions:}

1) To our best knowledge, this was the first (at least at its debut in arxiv) ternary weight quantization scheme to reduce storage and computational cost for deep neural networks.  


2) We propose an approximated and universal solution with threshold-based ternary function for calculating the ternary weights of the raw neural networks.

3) Experiments show the efficacy of our approach on public benchmarks for both image classification and detection.

\section{Ternary Weight Networks}
\vspace{-5pt}
\subsection{Advantage Overview}
\vspace{-5pt}
We address the limited storage and computational resources issues by introducing ternary weight networks (TWNs), which constrain the weights to be ternary-valued: +1, 0 and -1. TWNs seek to make a balance between the full precision weight networks (FPWNs) counterparts and the binary precision weight networks (BPWNs) counterparts. The detailed features are listed as follows.

\textbf{Expressive ability} In most recent network architectures such as VGG~\cite{Simonyan2014}, GoogLeNet~\cite{Szegedy2015} and ResNet~\cite{He2015}, a most commonly used convolutional filter is of size 3$\times$3. With binary precision, there is only $2^{3\times 3}$ = 512 templates. However, a ternary filter with the same size owns $3^{3\times 3}$ = 19683 templates, which gains 38$\times$ more stronger expressive abilities than the binary counterpart.

\textbf{Model compression} In TWNs, 2-bit storage requirement is needed for a unit of weight. Thus, TWNs achieve up to 16$\times$ model compression rate compared with the float32 precision counterparts. Take VGG-19~\cite{Simonyan2014} as an example, float version of the model needs $\sim$500M storage requirement, which can be reduced to $\sim$32M with ternary precision. Thus, although the compression rate of TWNs is 2$\times$ less than that of BPWNs, it is fair enough for compressing most of the existing state-of-the-art DNN models.

\textbf{Computational requirement} Compared with the BPWNs, TWNs own an extra zero state. However, the zero terms need not be accumulated for any multiple operations. Thus, the multiply-accumulate operations in TWNs keep unchanged compared with binary precision counterparts. As a result, it is also hardware-friendly for training large-scale networks with specialized DL hardware.

In the following parts, we will give detailed descriptions about the ternary weight networks problem and an approximated but efficient solution. After that, a simple training algorithm with error back-propagation is introduced and the run time usage is described at last.

\subsection{Problem Formulation}\vspace{-5pt}
To make the ternary weight networks perform well, we seek to minimize the Euclidian distance between the full precision weights $\mathbf{W}$ and the ternary-valued weights $\tilde{\mathbf{W}}$ along with a nonnegative scaling factor $\alpha$~\cite{Rastegari2016}. The optimization problem is formulated as follows,
\begin{equation}\label{eq1}
\left\{\begin{matrix} 
  \alpha ^*, \tilde{\mathbf{{W}^{*}}}= \underset{\alpha, \tilde{\mathbf{W}}}{\arg\min}\quad\mathbf{J}(\alpha, \tilde{\mathbf{W}}) =\left \| \mathbf{W-\alpha\tilde{\mathbf{W}}}  \right \|_{2}^{2}  \\  
  s.t.\quad\alpha \ge 0,\quad \tilde{\mathbf{W}_{i}}\in \left \{-1,0,+1  \right \},i=1,2...,n 
\end{matrix}\right.
\end{equation}

Here $\mathit{n}$ is the number of the filter. With the approximation $\mathbf{W}\approx \alpha\tilde{\mathbf{W}}$, a basic block of forward propagation in ternary weight networks is as follows,
\begin{equation}\label{eq2}
\left\{\begin{matrix} 
  \mathbf{Z}= \mathbf{X}\ast\mathbf{W}\approx \mathbf{X}\ast(\alpha\tilde{\mathbf{W}})=(\alpha\mathbf{X})\oplus\tilde{\mathbf{W}}   \\  
  \mathbf{X}^{next} = \mathit{g}(\mathbf{Z})\quad\quad\quad\quad\quad\quad\quad\quad\quad\quad\quad\quad\quad
\end{matrix}\right.
\end{equation}
where $\mathbf{X}$ is the input of the block; $\ast$ is a convolution or inner product operation; $\mathit{g}$ is a nonlinear activation function; $\oplus$ indicates a convolution or an inner product operation without multiplication; $\mathbf{Z}$ is the output feature map of the neural network block. It can also be used as input of the next block.

\subsection{Threshold-based Ternary Function}
\vspace{-5pt}

One way to solve the optimization Eq.~\ref{eq1}  is to expand the cost function $J(\alpha, \tilde{\mathbf{W}})$ and take the derivative w.r.t. $\alpha$ and $\tilde{\mathbf{W}_{i}}$ is respectively. However, this would get interdependent $\alpha^{*}$ and $\tilde{\mathbf{W}}_{i}^{*}$. Thus, there is no deterministic solution in this way~\cite{Hwang2014}. To overcome this, we try to find an approximated optimal solution with a threshold-based ternary function,
\begin{equation}\label{eq3}
  \tilde{\mathbf{W}_{i}} = f(\mathbf{W}_{i}|\Delta ) = \left\{\begin{matrix} 
  +1&\text{if}&\mathbf{W}_{i}>\Delta \\ 
  0&\text{if}&\left | \mathbf{W}_{i}\right |\le\Delta \\ 
  -1&\text{if}&\mathbf{W}_{i}<-\Delta 
\end{matrix}\right.  
\end{equation}

Here $\Delta$ is an positive threshold parameter. With Eq.~\ref{eq3}, the original problem can be transformed to
\begin{equation}\label{eq4}
    \alpha ^{*},\Delta^{*}=\underset{\alpha\ge 0,\Delta>0}{\arg\min} (|\mathbf{W}_{\Delta}|\alpha^{2}-2(\sum_{i\in \mathbf{I}_{\Delta}}|\mathbf{W}_{i}|)\alpha+c_{\Delta})
\end{equation}
where $\mathbf{I}_{\Delta}=\left \{ i{\Huge \vert}   \left | \mathbf{W}_{i}  \right | > \Delta   \right \} $ and $|\mathbf{I}_{\Delta}|$ denotes the number of elements in $\mathbf{I}_{\Delta}$; $c_{\Delta}={\textstyle \sum_{i\in \mathbf{I}_{\Delta }^{c}}^{}} $; $\mathbf{W}_{i}^{2}$ is a $\alpha$ independent constant. Thus, for any given $\Delta$, the optimal $\alpha$ can be computed as follows,
\begin{equation}\label{eq5}
    \alpha_{\Delta}^{*}=\frac{1}{\left | \mathbf{I}_{\Delta}  \right | } \sum_{i\in \mathbf{I}_{\Delta}}^{}\left | \mathbf{W}_{i}  \right |  
\end{equation}
By substituting $\alpha_{\Delta}^{*}$ into Eq.~\ref{eq4}, we get a $\Delta$ dependent equation, which can be simplified as follows,
\begin{equation}\label{eq6s}
\Delta^{*}=\underset{\Delta>0}{\arg\min}\frac{1}{\left | \mathbf{I}_{\Delta}  \right | }(\sum_{i\in \mathbf{I}_{\Delta}}^{}\left | \mathbf{W}_{i}  \right |  )^{2} 
\end{equation}

The above euqation has no straightforward solutions. Though discrete optimization can be made to solve the problem (due to states of $\mathbf{W}_{i}$ is finite), it should be very time consuming. As a viable alternative, we make a single assumption that $\mathbf{W}_{i}$ are generated from uniform or normal distribution. In case of $\mathbf{W}_{i}$ are uniformly distributed in $\left [ -\alpha ,\alpha  \right ] $ and $\Delta$ lies in $\left ( 0 ,\alpha  \right ] $, the approximated $\Delta^{*}$ is $\frac{\alpha }{3} $, which equals to $\frac{2}{3}\mathbf{E}(\left | \mathbf{W} \right | )$. When $\mathbf{W}_{i}$ is generated from normal distributions $N(0,\sigma ^{2})$, the approximated $\Delta^{*}$ is 0.6$\sigma$ which
equals to $0.75\mathbf{E}(\left | \mathbf{W}  \right | )$. Thus, we can use a rule of thumb that $\Delta^*\approx0.75\mathbf{E}(\left | \mathbf{W}  \right | ) \approx \frac{0.75}{n}  {\textstyle \sum_{i=1}^{n}}\left | \mathbf{W}_{i}  \right |$ for simplicity.

\begin{algorithm}[tb]
\caption{Train a M-layers CNN w/ ternary weights} \label{alg:twns}
\SetKwInOut{Input}{Inputs}
\Input{A minibatch of inputs and targets ($\bm{I}, \bm{Y}$), loss function $L(\bm{Y}, \hat{\bm{Y})}$ and current weight $\mathcal{W}^t$.}
\SetKwInOut{Input}{Hyper-parameter}
\Input{current learning rate $\eta^t$.}
\SetKwInOut{Output}{Outputs}
\Output{updated weight $\mathcal{W}^{t+1}$, updated learning rate $\eta^{t+1}$.}
Make the float32 weight filters as ternery ones:\\
\For{$m=~1~\text{to}~M$}{
    \For{$k^{th} \text{ filter in}~m^{th}~\text{layer}$}{
        $\Delta_{mk} = \frac{0.75}{n}\|\mathcal{W}^t_{mk}\|_{l1}$\\
        $\tilde{\mathcal{W}}_{mk}={\text{\{-1, 0, +1\}}}$ , refer to Eq.~\ref{eq3} \\
        $\alpha_{mk} = \frac{\mathcal{W}^t_{mk}\cdot\tilde{\mathcal{W}}_{mk}}{\tilde{\mathcal{W}}_{mk}\cdot\tilde{\mathcal{W}}_{mk}}$  \\
        $\mathcal{T}_{mk} = \alpha_{mk}\tilde{\mathcal{W}}_{mk}$\\
    }}
$\hat{\bm{Y}} = \text{\textbf{TernaryForward}}(\bm{I}, \tilde{\mathcal{W}}, \mathcal{\alpha})$ //\small{standard forward propagation} \\
$\frac{\partial L}{\partial \tilde{\mathcal{T}}} = \text{\textbf{TernaryBackward}}(\frac{\partial L}{\partial \hat{\bm{Y}}}, \tilde{\mathcal{T}})$ //\small{standard backward propagation except that gradients are computed using $\mathcal{T}$ instead of $\mathcal{W}^t$} \\
$\mathcal{W}^{t+1} = \text{\textbf{UpdateParameters}}(\mathcal{W}^t, \frac{\partial L}{\partial \mathcal{T}}, \eta_t)$ //\small{ we use SGD in this paper }\\
$\eta^{t+1} = \text{\textbf{UpdateLearningrate}}(\eta^t, t)$ //\small{we use learning rate step decay in this paper}
\end{algorithm}

\subsection{Training of Ternary-Weight-Networks}
\begin{figure}[tb!]
  \centering
  \includegraphics[width=0.48\textwidth]{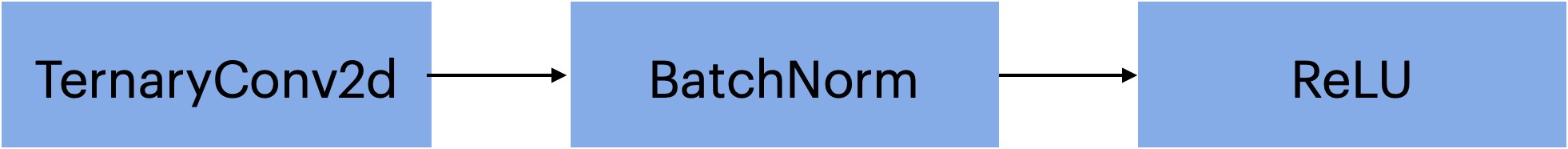}
  \caption{A typical Ternary block in TWNs. In the forward pass, we apply ternarization operation for the weight of  convolution layer meanwhile the float32 weight will be cached for future parameter update; in the backward pass, we calculate ternary weight gradient to update the float32 weight.}
  \label{fig:twnsconvblock}
\end{figure}

\vspace{-5pt}
CNNs typically includes Convolution layer, Fully-Connected layer, Pooling layer (e.g.,Max-Pooling, Avg-Pooling), Batch-Normalization (BN) layer~\cite{Ioffe2015} and Activation layer (e.g.,ReLU, Sigmoid), in TWNs, we also follow the traditional neural network block design philosophy, the order of layers in a typical ternary block of TWNs is shown in Fig.~\ref{fig:twnsconvblock}.  

We borrow the parameter optimization strategy which successfully applied from BinaryConncet~\cite{courbariaux2015binaryconnect} and XNOR-Net~\cite{Rastegari2016}, in our design, ternarization only happens at the forward and backward pass in convolution and fully-connected layers, but in the parameters update stage, we still keep a copy of the full-precision parameters. In addition, two effective tricks, Batch-Normalization and learning rate step decay that drops the learning rate by a factor every few epochs, are adopted. We use stochastic gradient descent (SGD) with momentum to update the the parameters when training TWNs, the detailed training strategy show in Table~\ref{table:expconf}. 

\subsection{Inference of Ternary-Weight-Networks}
\vspace{-5pt}

In the forward pass, the scaling factor $\alpha$ could be transformed to the inputs according to Eq.~\ref{eq2}.
Thus, we only need to keep the ternary-valued weights and the scaling factors for deployment. This
would results up to 16$\times$ model compression rate for deployment compared with the float32 precision counterparts.

\begin{figure*}[tb!]
	\centering
	\subfigure[MNIST]{
		\begin{minipage}[b]{0.3\textwidth}
	    \includegraphics[width=1\textwidth]{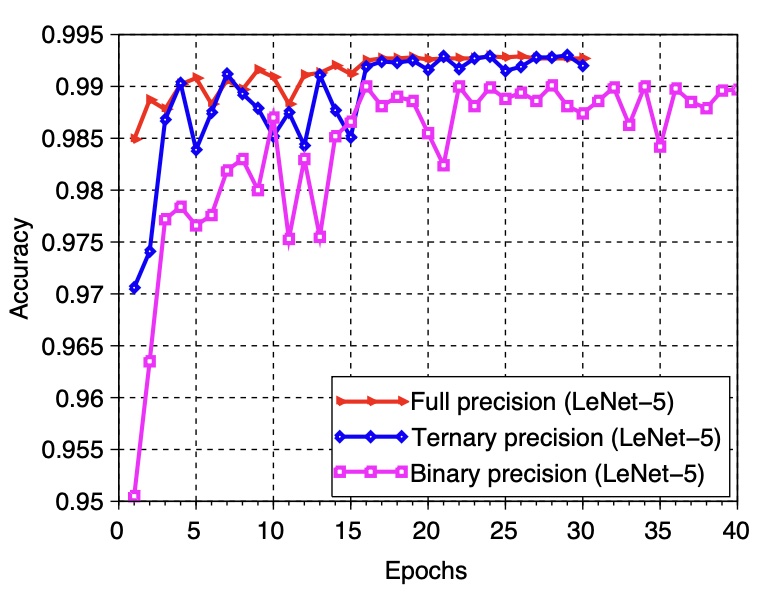}
		\end{minipage}
		\label{fig:hor_2figs_1cap_2subcap_1}
	}
    \subfigure[CIFAR-10]{
    	\begin{minipage}[b]{0.3\textwidth}
   		\includegraphics[width=1\textwidth]{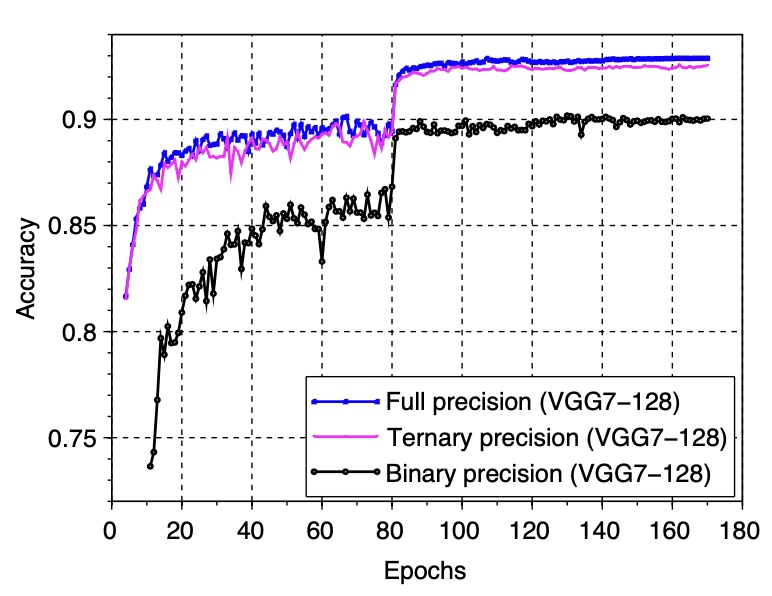}
    	\end{minipage}
		\label{fig:hor_2figs_1cap_2subcap_2}
    	}
    \subfigure[ImageNet (top-5)]{
    	\begin{minipage}[b]{0.3\textwidth}
   		\includegraphics[width=1\textwidth]{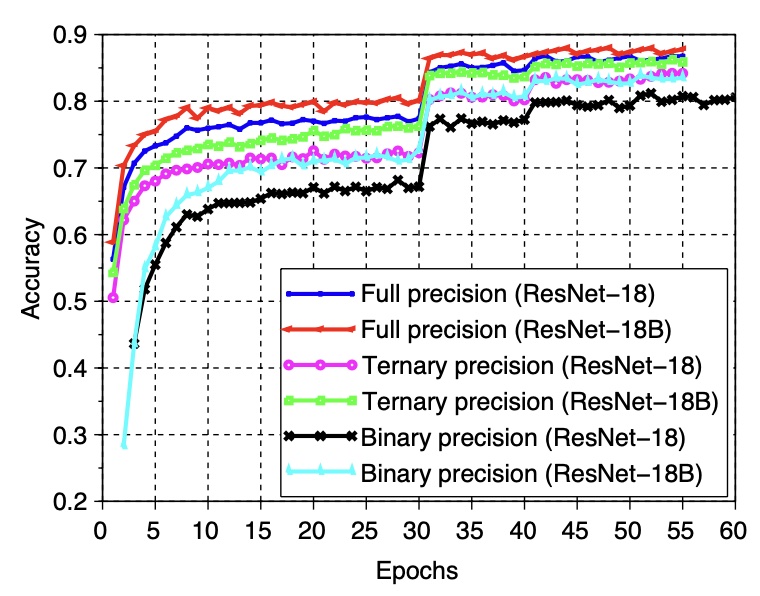}
    	\end{minipage}
		\label{fig:hor_2figs_1cap_2subcap_3}
    	}
    	\vspace{-10pt}
	\caption{Classification accuracy over training epochs MNIST (top-1 accuracy), CIFAR10 (top-1) and ImageNet (top-5).}
	\label{fig:benchresults}
\end{figure*}

\section{Experiments and Discussion}
\vspace{-5pt}
We benchmark Ternary Weight Networks (TWNs) with Binary Weight Networks (BPWNs) and Full Precision Networks (FPWNs) on both classification task (MNIST, CIFAR-10 and ImageNet) and object detection task (PASCAL VOC).

For a fair comparison, we keep the following configures to be same: network architecture, regularization method (L2 weight decay), learning rate scaling procedure (multi-step) and optimization method (SGD with momentum). BPWNs use sign function to binarize the weights and FPWNs use float-valued weights. See Table~\ref{table:expconf} for   training configurations.

\begin{table}[tb!]
    \centering
    \caption{Backbones and hyperparameters setting for different datasets used by our method on three benchmarks.}
    \begin{threeparttable}
\resizebox{.98\linewidth}{!}{
    \begin{tabular}{@{}cccc@{}}
\toprule
\multicolumn{1}{l}{}      & MNIST   & CIFAR-10 & ImageNet                         \\ \midrule
backbone architecture      & LeNet-5 & VGG-7    & \multicolumn{1}{l}{ResNet18B} \\
weight decay              & 1e-4    & 1e-4     & 1e-4                             \\
mini-batch size           & 50      & 100      & 64(x4)\tablefootnote{We use 4 GPUs to speed up the training.}                             \\
initial learning rate     & 0.01    & 0.1      & 0.1                              \\
learning rate adjust step\tablefootnote{Learning rate is divided by 10 at these epochs.} & 15, 25  & 80, 120  & 30, 40, 50                       \\
momentum                  & 0.9     & 0.9      & 0.9                              \\ \bottomrule
\end{tabular}}
    \end{threeparttable}
\label{table:expconf} 
\end{table}
\subsection{Experiments of Classification}\vspace{-5pt}
\textbf{MNIST} is a collection of handwritten digits. It is a very popular dataset in the field of image processing. The LeNet-5~\cite{LeCun1998} architecture we used in MNIST experiment is “32-C5 + MP2 + 64-C5 + MP2 + 512 FC + SVM” which starts with a 5x5 convolutional block that includes a convolution layer, a BN layer and a relu layer. Then a max-pooling layer is followed with stride 2. The “FC” is a fully connect block with 512 nodes. The top layer is a SVM classifier with 10 labels. Finally, hinge loss is minimized with SGD.

\textbf{CIFAR-10} consists of 10 classes with 6K color images of 32x32 resolution for each class. It is divided into 50K training and 10K testing images. We define a VGG inspired architecture, denoted as VGG-7, by ``2$\times$(128-C3) + MP2 + 2$\times$(256-C3) + MP2 + 2$\times$(512-C3) + MP2 + 1024-FC + Softmax”. Compared with the architecture in~\cite{courbariaux2015binaryconnect}, we ignore the last fully connected layer. We follow the data augmentation in~\cite{He2015,Lee2015} for training: 4 pixels are padded on each side, and a 32$\times$32 crop is randomly sampled from the padded image or its horizontal flip. At testing time, we only evaluate the single view of the original 32$\times$32 image.

\textbf{ImageNet} consists of about 1.2 million train images from 1000 categories and 50,000 validation images. ImageNet has higher resolution and greater diversity, is more close to real life than MNIST and CIFAR-10. We adopt the popular ResNet18 architecture~\cite{He2015} as backbone. Besides, we also benchmark another enlarged counterpart whose number of filters in each block is 1.5$\times$ of the original one {which is termed as ResNet18B}. In each training iteration, images are randomly cropped with 224$\times$224 size. We do not use any resize tricks~\cite{Rastegari2016} or any color augmentation. 

\begin{table*}[tb!]
    \centering
    \caption{Classification accuracy (\%) on ImageNet with ResNet18 (or ResNet18B in bracket) as backbones.}
\label{table:validacc}
\begin{tabular}{ccccc} 
\hline
\multicolumn{1}{l}{}      & MNIST & CIFAR-10 & ImageNet (top-1) & \multicolumn{1}{l}{ImageNet (top-5)}  \\ 
\midrule
TWNs (our main approach)                      &99.35 & 92.56    &61.80 (65.3)     &84.20 (86.2)                          \\
BPWNs (binary precision counterpart)                     & 99.05 & 90.18    & 57.50 (61.6)     & 81.20 (83.9)                          \\ 
\cline{2-5}
FPWNs (full precision counterpart)                          & 99.41 & 92.88    & 65.4 (67.6)     & 86.76 (88.0)                         \\ 
\hline
BinaryConnect~\cite{courbariaux2015binaryconnect}             & 98.82 & 91.73    & -               & -                                    \\
Binarized Neural Networks~\cite{Hubara2016} & 98.60  & 89.85    & -               & -                                    \\
Binary Weight Networks~\cite{Rastegari2016}    & -     & -        & 60.8            & 83.0                                 \\
XNOR-Net~\cite{Rastegari2016}                  & -     & -        & 51.2            & 73.2                                 \\
\hline
\end{tabular}
\end{table*}

\begin{table*}[tb!]
    \centering
    \caption{Detection performance (\%) on PASCAL VOC with YOLOv5 (small) as detector on Pascal VOC.}
\label{table:voc}
\begin{tabular}{ccccc} 
\hline
\multicolumn{1}{l}{}& Precision & Recall & mAP\_{50} & mAP\_{50:95}  \\ 
\midrule
TWNs (our main approach)                 &  78.0\%   & 69.1\% & 76.8\% & 51.5\%     \\
BPWNs (binary precision counterpart)                    & 69.8\% & 56.7\% & 62.9\% & 39.4\%  \\ 
FPWNs (full precision counterpart)                        & 83.3\% & 80.8\% & 86.7\% & 63.7\%                      \\ 
\hline
\end{tabular}
\end{table*}

Table~\ref{table:validacc} shows the classification results. On the small datasets (MNIST and CIFAR-10), TWNs achieve similar performance as FPWNs, while beats BPWNs. On the large-scale ImageNet dataset, BPWNs and TWNs both get poorer performance than FPWNs. However, the accuracy gap between TWNs and FPWNs is smaller than the gap between BPWNs and TWNs. In addition, when we change the backbone from ResNet18 to ResNet18B, as the model size is larger, the performance gap between TWNs (or BPWNs) and FPWNs has been reduced. This indicates low precision networks gain more merits from larger models than the full precision counterparts.

The validation  accuracy curves of different approaches across all training epochs on MNIST, CIFAR-10 and ImageNet datasets illustrate in Fig.~\ref{fig:benchresults}. As we can see in the figure, obviously, BPWNs converge slowly and the training loss is not stable compared with TWNs and FPWNs. However, TWNs converge almost as fast and stably as FPWNs.

\subsection{Experiments of Detection}\vspace{-5pt}
\textbf{PASCAL VOC}~\cite{voc} consists of 20 classes with 11540 images and 27450 labeled objects. We adopt the popular YOLOv5 (small)~\cite{yolov5} architecture and compare the performance of full precision, binary precision  and ternary precision in Table~\ref{table:voc}. Specifically, we initialize each model by the weights trained on MS-COCO dataset~\cite{coco} (provided by YOLOv5) and fine-tune each model by 150 epochs. We observe that TWNs significantly outperforms BPWNs by more than 10\% mAP, showing the great effectiveness of our method.

\section{Conclusion}
\vspace{-5pt}
In this paper, we have introduced the simple, efficient, and accurate ternary weight networks for real world AI application which can reduce the memory usage about 16x and the the computation about 2x. We present the optimization problem of TWNs and give an approximated solution with a simple but effective ternary function. The proposed TWNs achieve a balance between accuracy and model compression rate as well as potentially low computational requirements of BPWNs. Empirical results on public benchmarks show the superior performance of the proposed method.

\bibliographystyle{IEEEbib}
\bibliography{icassp}

\end{document}